\documentclass[pdflatex,sn-mathphys-num]{sn-jnl}

\usepackage{graphicx}%
\usepackage{multirow}%
\usepackage{amsmath,amssymb,amsfonts}%
\usepackage{amsthm}%
\usepackage{mathrsfs}%
\usepackage[title]{appendix}%
\usepackage{xcolor}%
\usepackage{textcomp}%
\usepackage{manyfoot}%
\usepackage{booktabs}%
\usepackage{algorithm}%
\usepackage{algorithmicx}%
\usepackage{algpseudocode}%
\usepackage{listings}%

\usepackage{tikz}
\usetikzlibrary{matrix,shapes.geometric, arrows, positioning, fit}
\usepackage[strings]{underscore}
\usepackage{tabularx,ragged2e}
\newcolumntype{L}{>{\RaggedRight}X}

\theoremstyle{thmstyleone}%
\theoremstyle{thmstyletwo}%

\theoremstyle{thmstylethree}%

\begin{document}

\title[Article Title]{Unboxing the Black Box: Mechanistic Interpretability for Algorithmic Understanding of Neural Networks}


\author*[1]{\fnm{Bianka} \sur{Kowalska}}\email{bianka.kowalska@pwr.edu.pl}
\author[1]{\fnm{Halina} \sur{Kwaśnicka}}\email{halina.kwasnicka@pwr.edu.pl}

\affil*[1]{\orgdiv{Department of Artificial Intelligence}, \orgname{Wroclaw University of Science and Technology}, \orgaddress{\street{Wybrzeże Stanisława Wyspiańskiego 27}, \city{Wrocław}, \postcode{50-370}, \country{Poland}}}


\abstract{The black box nature of deep neural networks poses a significant challenge for the deployment of transparent and trustworthy artificial intelligence (AI) systems. 
With the growing presence of AI in society, it becomes increasingly important to develop methods that can explain and interpret the decisions made by these systems. 
To address this, mechanistic interpretability (MI) emerged as a promising and distinctive research program within the broader field of explainable artificial intelligence (XAI).
MI is the process of studying the inner computations of neural networks and translating them into human-understandable algorithms.
It encompasses reverse engineering techniques aimed at uncovering the computational algorithms implemented by neural networks.
In this article, we propose a unified taxonomy of MI approaches and provide a detailed analysis of key techniques, illustrated with concrete examples and pseudo-code. We contextualize MI within the broader interpretability landscape, comparing its goals, methods, and insights to other strands of XAI. 
Additionally, we trace the development of MI as a research area, highlighting its conceptual roots and the accelerating pace of recent work. We argue that MI holds significant potential to support a more scientific understanding of machine learning systems -- treating models not only as tools for solving tasks, but also as systems to be studied and understood. We hope to invite new researchers into the field of mechanistic interpretability.}




\maketitle

\section{Introduction}

Artificial intelligence (AI) is increasingly assisting us in a wide range of tasks, from everyday applications like recommendation systems to high-risk domains such as biometric recognition, autonomous vehicles, and medical diagnosis~\cite{ali_explainable_2023}. In particular, the rise of transformer-based models, such as those used in natural language processing (NLP), has significantly accelerated AI's adoption and visibility in society, enabling breakthroughs in fields like text generation, translation, and image understanding~\cite{lin_survey_2022}. The size, complexity, and opacity of deep learning models are growing exponentially, further outpacing the ability of researchers to understand the black box. As deep neural networks are increasingly deployed in real-world applications with more advanced use cases, the impact of AI continues to grow. This growing influence, coupled with the often opaque, black-box nature of most AI systems, has led to a heightened demand for AI models that are both faithful and explainable. The validation of AI's decisions is especially critical in high-risks areas, such as law or medicine~\cite{barredo_arrieta_explainable_2020,srinivasu_blackbox_2022}. 
As a result, Explainable AI (XAI) emerged as a direct response to companies' and researchers' demands to interpret, explain and validate neural networks to make AI systems trustworthy.

XAI encompasses all methods, approaches and efforts to uncover the reasoning and behavior of artificial intelligence systems~\cite{ali_explainable_2023}. Thus, it is important to establish an understanding of common terms used in the XAI literature, despite the lack of universally accepted definitions. In general, \emph{explainability} refers to the understanding of system's outputs~\cite{hamida_exploring_2024}, answering the \emph{why?} question; whereas interpretability refers to the understanding of the system's internal workings~\cite{das_opportunities_2020,rauker_toward_2023}, answering the \emph{how?} question.  
\textit{Interpretable} artificial intelligence refers to AI models that are intrinsically transparent in their operation, thereby eliminating the need for post-hoc analysis. Consequently, these models enable the identification and tracking of causal relationships~\cite{schwalbe_comprehensive_2024}. \textit{Trustworthy} AI is mainly associated with AI systems designed to inspire trust by prioritizing safety, fairness, and transparency. The employment of XAI is a viable method for the creation of trustworthy AI models~\cite{kowald_establishing_2024,markus_role_2021}. Table~\ref{tab:xaitai} summarizes the key aspects of XAI.

\begin{table}[htbp]

\small
\caption{Different aspects of XAI}
\label{tab:xaitai}

\begin{tabularx}{\textwidth}{@{}lLLL@{}}
\toprule
\textbf{Aspect} & \textbf{Explainable AI} & \textbf{Interpretable AI} & \textbf{Trustworthy AI} \\
\midrule
Focus&Post-hoc explanations of models' decisions& Transparent models by design & Safe and ethical AI systems\\
Goals&Explaining the model’s decision&Understanding of models' functioning&Building trust in AI systems\\
Examples&Feature importance&Decision trees&Bias-free hiring system\\
\botrule
\end{tabularx}
\end{table}

\begin{figure}[h]
\centering
\includegraphics[width=0.9\textwidth]{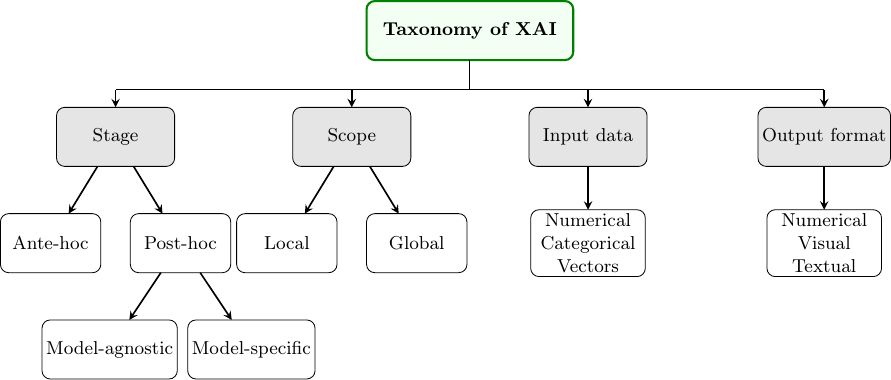}
\caption{Brief taxonomy of XAI methods}\label{xai_taxonomy}
\end{figure}

The growing demand for transparency in AI systems has led to the development of a vast number of Explainable AI methods, each designed to address different aspects of interpretability across various models and applications. Researchers have proposed different taxonomies to address this diversity~\cite{a_systematic_2023,ali_explainable_2023,barredo_arrieta_explainable_2020,schwalbe_comprehensive_2024}. 
The key distinctions between XAI methods are scope (local explanations for individual predictions versus global insights into overall model behavior) and stage of application (whether explanations are integrated into the model or applied post-hoc). Furthermore, post-hoc approaches are divided into model-specific methods for explaining a certain model and model-agnostic methods working independently of the underlying model~\cite{speith_review_2022}. Other distinctions may include the output format of explanations, the model's input data or problem type. Figure \ref{xai_taxonomy} presents the key aspects in the taxonomy of XAI methods.

One notable approach that emerged in this decade is Mechanistic Interpretability (MI), a cognition-based set of reverse engineering techniques aimed at uncovering the computational algorithms of neural networks~\cite{olah_mechanistic_2022,saphra_mechanistic_2024}. 
Mechanistic interpretability focuses on the internal structures of AI models -- neurons, layers, attention heads, and circuits -- and their causal roles in computation. As opposed to explainability methods, such as feature attribution (e.g., LIME~\cite{ribeiro_why_2016} or SHAP~\cite{lundberg_unified_2017}), MI seeks a deeper understanding of models, drawing inspiration from neuroscience and systems biology. According to XAI taxonomy, it can be categorized as a post-hoc, model-specific approach.
Rather than treating an AI system as a black-box, MI researchers emphasize the importance of inner interpretability based on the premise that internal components of neural networks adopt specific roles after training~\cite{manning_emergent_2020,saphra_training_2021}. Identification of such components provides deeper explanations and a scientific understanding of AI. To uncover these roles, researchers employ a range of techniques rooted in pattern recognition, as any non-random structure within a neural network can serve as a functional component~\cite{kastner_explaining_2024}.
These techniques include methods for analyzing the representations learned by models, assessing the causal impact of specific components, and visualizing patterns within the network's structure and activations.  

Mechanistic interpretability allows for an organized characterization of AI systems, as opposed to the divide-and-conquer methods of XAI which provide explainability only in specific contexts~\cite{kastner_explaining_2024}. Its primary goal is to enhance AI safety and foster trust by delivering comprehensive, human-interpretable explanations of deep models. This is achieved by mapping low-level components, such as circuits or neurons, to elements of human-understandable algorithms and progressively building upward to construct a comprehensive understanding~\cite{olah_mechanistic_2022}. 
As research in this area progresses, research efforts are increasingly focused on refining methodologies and scaling analyses to more complex models.
The increasing interest in mechanistic interpretability is further evidenced by the growing number of publications, particularly review articles. Kästner and Crook~\cite{kastner_explaining_2024}, Grzankowski~\cite{grzankowski_real_2024}, and Rabiza~\cite{rabiza_mechanistic_2024} situate mechanistic interpretability in philosophical context. Davies and Khakzar~\cite{davies_cognitive_2024} seek parallels between MI and cognitive science. Bereska and Gavves~\cite{bereska_mechanistic_2024} present a comprehensive overview of the current advancements in MI, as well as a unified taxonomy and explanation of different concepts, all within the overarching context of ensuring safety. Rai et al.~\cite{rai_practical_2025} focus on employing MI in language models, and Saphra and Wiegreffe~\cite{saphra_mechanistic_2024} present the divide between MI and NLP interpretability communities. Sharkey et al.~\cite{sharkey_open_2025} focus on open challenges in the field and present possible future directions.

The focus of this article is to identify the current status of mechanistic interpretability. We position MI within the broader XAI community and propose a clear definition, along with a taxonomy of techniques. We also present the research background of MI -- by providing a comprehensive overview of MI approaches and tasks, alongside multiple examples of articles and research in this field. Specifically, we pursue the following objectives:
\begin{itemize}
    \item Situating mechanistic interpretability within the broader context of mechanistic explanations and explainable AI (Section~\ref{what_is_mi}).
    \item Providing a clear definition and taxonomy of mechanistic interpretability (Section~\ref{what_is_mi}).
    \item Conducting a literature review on mechanistic interpretability (Section~\ref{literature}). 
    \item Presenting the key techniques used in the field, with pseudo-code and literature examples (Section~\ref{methods}).
    \item Outlining major challenges and opportunities in MI (Section~\ref{future}).
\end{itemize}

The paper is structured as follows: Section~\ref{what_is_mi} describes the philosophical background of MI and places it in the broader XAI community. It follows with a brief introduction to the core concepts in MI.
Section~\ref{literature} presents a literature review. Section~\ref{methods} describes the different approaches used in MI and presents notable advancements. Section~\ref{future} addresses the challenges and opportunities found in MI. Section~\ref{conclusion} outlines future goals in MI and concludes the article.

\section{What is Mechanistic Interpretability?} \label{what_is_mi}

The mechanistic approach to understanding neural networks is a relatively new development within the field of AI interpretability. As this approach gains traction, it is important to situate it within the broader XAI community. In this section, we first outline the scientific and historical context of MI, followed by the presentation of a definition and taxonomy.

\subsection{Origins of Mechanistic Interpretability} 

\textbf{Philosophical background.} The study of mechanisms and mechanistic explanations is rooted in science and philosophy and has been known since the 20th century~\cite{craver_mechanisms_2024}. Bechtel and Abrahamsen~\cite{bechtel_explanation_2005} define a mechanism as a \emph{structure performing a function in virtue of its components parts, component operations, and their organization}. In other words, a mechanism is a connected structure that causes some phenomena through interaction between parts, and can be represented as a directed graph of components involved in the process. By studying the phenomenon, researchers aim to uncover the underlying mechanism and explain its functioning in terms of its causal components. This investigative process culminates in the formulation of a mechanistic explanation~\cite{machamer_thinking_2000}.
In fields like medicine~\cite{courcelles_solving_2022,metzcar_review_2024}, physics, and neuroscience~\cite{bassett_network_2017}, mechanistic explanations play a critical role in dissecting complex systems, revealing their constituent parts, and establishing causal relationships that elucidate how these components work together to produce emergent behavior. 

Mechanistic interpretability parallels the way scientists study physical systems and aligns AI research with the foundational principles of explanatory science. It conceptualizes artificial neural networks as complex systems composed of multiple mechanisms, establishing an analogy between AI and the human brain~\cite{davies_cognitive_2024,liu_seeing_2023,fernando_transformer_2025,mineault_neuroai_2024}. This approach is further supported by the past applications of mechanistic explanations in neuroscience and cognitive sciences~\cite{craver_explaining_2007,craver_mechanisms_2024,kostic_mapping_2023,khambhati_modeling_2018}. By bridging insights from neuroscience and AI, mechanistic interpretability offers a new framework for understanding artificial networks. \newline

\noindent\textbf{XAI background.} Mechanistic interpretability was first brought up in the context of interpreting vision models~\cite{olah_zoom_2020}. With most of its early research being published on blogs and forums (see, e.g., \href{https://www.lesswrong.com/tag/interpretability-ml-and-ai}{LessWrong}), the field developed separately from other interpretability methods. 
At the same time, due to the rise of advanced transformer models, much attention was given in the broader XAI community to the explanations of large language models. As a result, a NLP interpretability community unfolded, which at the time was mostly unaware of the mechanistic research. Concurrently, a great proportion of MI researchers have turned their attention to interpreting transformers~\cite{elhage_mathematical_2021}. Thus, two distinct communities emerged, united by the same goal: AI safety~\cite{saphra_mechanistic_2024}.

When MI researchers engaged with academia, a clash with the NLP interpretability community emerged. Initially, many NLP researchers criticized MI research for "rediscovering" existing interpretability methods that had already been adopted and refined within the NLP community. This tension was fueled by differing terminologies and methodological approaches, leading to debates over the novelty and relevance of MI contributions. NLP researchers argued that MI often overlooked the rich history and established practices of NLP interpretability, while MI proponents contended that their focus on detailed, mechanistic explanations provided deeper insights into model behavior. Despite these conflicts, the clash ultimately spurred a productive dialogue, encouraging both communities to re-evaluate and integrate their methods, fostering a more comprehensive understanding of AI interpretability.

Although both mechanistic and NLP interpretability communities work towards similar goals, the term "mechanistic" has not been broadly adapted in the academia. Interpretability methods can both be described as mechanistic and not, leading to a blurred line between taxonomies and categorizations of approaches -- for instance probing, which was present long before MI, has been adopted by MI researchers and is now often classified as a mechanistic method~\cite{alain_understanding_2017,gurnee_finding_2023}. This overlap has led to a rich, albeit complex, landscape of interpretability research, where methods and terminologies often intersect and evolve.

\subsection{Defining Mechanistic Interpretability}
Following on the problems described in previous section, we propose a new definition for mechanistic interpretability: one that joins MI pursuit with the overall \textbf{inner interpretability} research, but distinguishes it by focusing on the computational aspects of neural networks. 

\begin{sloppypar}
\textbf{Definition.} \textit{Mechanistic interpretability is the process of studying the inner computations of neural networks and translating them into human-understandable algorithms.}
\end{sloppypar}

This definition supports the core idea of MI, which is its algorithmic approach to neural networks. Olah~\cite{olah_mechanistic_2022} compares MI to reverse-engineering a compiled computer program. Similarly to variables interacting in programs to produce a desired output, neural networks are composed of neurons, interacting through their connections via weights. Based on the premise that neurons can be understood, and therefore the connections between them, MI aims to deconstruct neural networks into interpretable subnetworks. Although a neural network as a whole is a complex structure, it can be broken down into small, meaningful components. 

The fundamental unit of neural network representations is a \textbf{feature}. Features refer to any interpretable properties or characteristics encoded in neurons. 
Unlike traditional machine learning, where features correspond directly to input data attributes~\cite{bishop_pattern_2006}, MI treats them as abstractions capturing internal model properties.
For instance, a feature in a vision model could be a curve detector, which responds to images containing a curve~\cite{olah_zoom_2020}. Formally, they are defined as directions in the network's activation space~\cite{elhage_toy_2022}. Ideally, a single feature would correspond to a single neuron. However, in real-world neural networks a major challenge is \textbf{polysemanticity} — the phenomenon where a single neuron encodes multiple features to accommodate the model's limited number of neurons~\cite{olah_zoom_2020,elhage_toy_2022}. A polysemantic neuron responds to different, often unrelated concepts, undermining the mechanistic approach to AI. Polysemanticity is explored in detail in Section~\ref{disentanglement}.

MI seeks to interpret the connections and interactions between features by analyzing \textbf{circuits} -- features linked by weights. Circuits are the minimal computational subgraphs of neural networks that are responsible for performing a given task. They are defined as directed, acyclic graphs. For instance, the curve detector feature is linked with consequent neurons to represent more complex features~\cite{olah_zoom_2020}. Formally, in a neural network represented as a directed graph $G=(V,E)$, where $V$ is the set of nodes (neurons in the network) and $E$ is the set of weighted edges (representing weighted connections between neurons), a circuit is a subgraph within $G$: $C=(V_C,E_C)$, where $V_C \subseteq V$ and $E_C \subseteq E$, that encapsulates a coherent computational process at various levels of abstraction. 

Circuit discovery is the leading task defining mechanistic interpretability, showing promising results in both smaller and more complex models~\cite{lieberum_does_2023}. The study of circuits could pave the way for proving the universality hypothesis -- the idea that similar circuits and features are present across tasks and models. Such reoccurring patterns are \textbf{motifs}. The universality hypothesis, if true, may allow for knowledge transfer from small, toy models to state-of-the-art large-scale models~\cite{chughtai_toy_2023,merullo_circuit_2024,nainani_adaptive_2024}. \newline

\begin{figure}[h]
\centering
\includegraphics{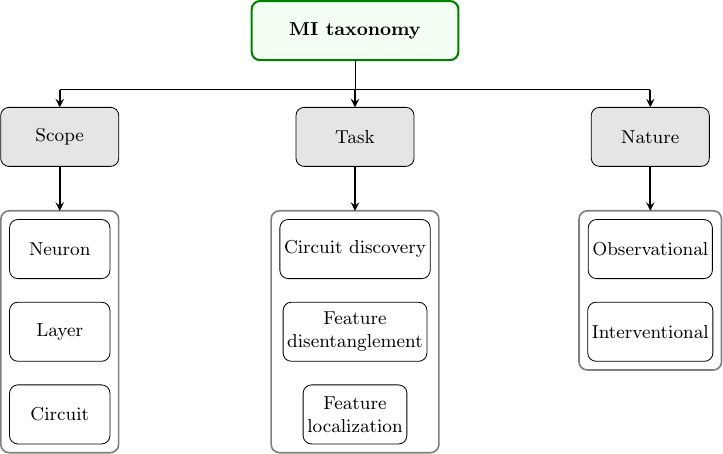}
\caption{Taxonomy of MI methods. The component and task criteria are not limited to the presented examples}\label{mi_taxonomy}
\end{figure}

\noindent\textbf{Taxonomy of MI approaches.}
Mechanistic interpretability approaches can be organized based on different levels of analysis. Although all methods are united by the "top-down approach" aimed at unfolding input-output mechanisms responsible for a model's behaviors, they employ different techniques, and the definition of the entities analyzed varies across studies.  
The proposed taxonomy divides MI approaches using three criteria: scope, task, and nature of analysis (Figure~\ref{mi_taxonomy}). 

\noindent\textit{Scope.} Scope refers to the part of neural network being interpreted. Any neural network component involved in the forward pass can be interpreted as a mechanism entity~\cite{rai_practical_2025}. A narrow analysis focuses on individual neurons and identifying their role in specific tasks; for instance, identifying a neuron crucial for a given decision. Over time, researchers have been generalizing to studying more complex elements, such as attention heads or whole layers~\cite{wang_interpretability_2022}. Broader scope, such as circuit, refers to studying structures within a network, where the connections between components play a crucial role in interpretation.

\noindent\textit{Task.} The primary objective of MI is to reverse-engineer a neural network; however, most of current research addresses only fragments of this goal. Among these are tasks such as feature localization, circuit discovery, and feature disentanglement. Feature localization refers to identifying components that are responsible for specific features. Circuit discovery is the task of uncovering meaningful structures in the network. Feature disentanglement focuses on resolving polysemanticity. While all contribute valuable pieces to the puzzle, they fall short of providing a holistic understanding of neural networks. researchers often combine multiple methods in order to achieve an understanding (e.g., validating a circuit found with patching via probing or visualization~\cite{marks_geometry_2024}).

\noindent\textit{Nature of analysis.} MI can be divided into two main categories: observation-based approaches and intervention-based approaches. The key difference lies in whether the focus is on passively observing the model's behavior or actively modifying its components to study the effects.
Observation-based approaches concentrate on examining the internal mechanisms of a neural network without making any changes to its architecture or parameters. These methods typically involve techniques such as visualizing weights, activations, or gradients, as well as tracing the flow of information through the network. Conversely, intervention-based approaches involve actively altering the network or its inputs to investigate causal relationships and gain insights into specific behaviors. 

%
\section{The Rise of Mechanistic Interpretability} \label{literature}
We present a comprehensive survey of literature to illustrate the rapid emergence and growth of mechanistic interpretability, highlighting key trends and the increasing volume of research in recent years. All search phrases used in the search were enclosed in quotation marks. \newline

%

\noindent\textbf{Methodology.} 
At the start of our research, we searched roughly the \textit{Web of Science} database without time restrictions. We aimed to present MI within the broader context of research on AI explanation and interpretability. Since publications specifically focused on the mechanistic interpretability of AI models only began to appear around 2015, we examined the wider field using the following keywords:
\begin{itemize}
    \item \textit{Explainable Artificial Intelligence} (similar results with \textit{Explainable Artificial Intelligence OR Explainable Machine Learning}): One work appeared in 2004, 2005 and 2006; two in 2017, and 22 in 2018, followed by a rapid growth.
    \item \textit{Interpretable Artificial Intelligence OR Interpretable Machine Learning}: The first work appeared in 2006; growth remained slow until 2018 (33 works), after which it accelerated. For \textit{Interpretable Machine Learning} alone, the number rose from 4 works in 2014 to 960 in 2024.
   \item \textit{Mechanistic Interpretability AND Machine Learning}: isolated works appeared between 2010 and 2022; In 2024, 23 publications were listed.
\end{itemize}
Based on this and additional database analyses, we decided to focus our quantitative analysis on the period since 2015, reflecting the dynamic growth of AI model interpretation methods.
%

The literature search was conducted in three databases: \textit{Scopus}~\footnote{\url{https://www.scopus.com}}, \textit{Web of Science}~\footnote{\url{www.webofscience.com/wos/}}, and \textit{PubMed}~\footnote{\url{https://pubmed.ncbi.nlm.nih.gov}}. 
We encountered two major challenges when researching the mechanistic interpretability of neural networks:
\begin{itemize}
    \item \textit{Exclusion}: concerning papers related to mechanistic interpretability, but not mentioning their mechanistic background directly. 
    \item \textit{Broadness}: concerning papers related to mechanistic explanations in general but not to AI models.
\end{itemize}

\textit{Exclusion} is caused by the novelty of MI in the context of AI interpretability. 
Many researchers who study topics related to mechanistic interpretability may not be aware of this naming convention. Moreover, as exemplified with probing, some approaches are now considered mechanistic despite their different background. In addition, much of the pursuit of MI has been published in blogs and forums, thus it is not listed in research databases.

\textit{Broadness} is attributable to the philosophical and scientific background of MI. Mechanistic explanations are widely used across various disciplines, including biology, physics, and engineering.
This broad applicability makes it difficult to filter and curate AI-specific literature, as many papers discuss mechanistic insights without being directly relevant to neural network interpretability. Thus, the search term "mechanistic interpretability" often yields results not related to neural networks interpretability. 

We studied several search terms, including "mechanistic interpretability" which suffered from broadness, "mechanistic interpretability AND neural networks", which suffered from exclusion. 
Ultimately, we limited the scope of the search to "Mechanistic interpretability AND Machine Learning". This combination both encompassed the literature related to interpretability of deep neural networks and excluded works related to mechanistic explanations in other disciplines. 

The problem of broadness was especially prevalent in PubMed. The majority of papers were related to mechanistic explanations in medicine, like for instance~\cite{gaw_integration_2019}, where the authors combine machine learning with mechanistic models to study glioblastoma. Distinguishing AI-focused mechanistic interpretability papers from the broader landscape of mechanistic explanations would require careful manual selection.
Conversely, results from Web of Science suffered from exclusion and yielded little results when searching for "Mechanistic interpretability" (24 papers on April 17th). The results were accurate, yet sparse due to exclusion. \newline

\begin{figure}[h]
\centering
\includegraphics[width=0.9\textwidth]{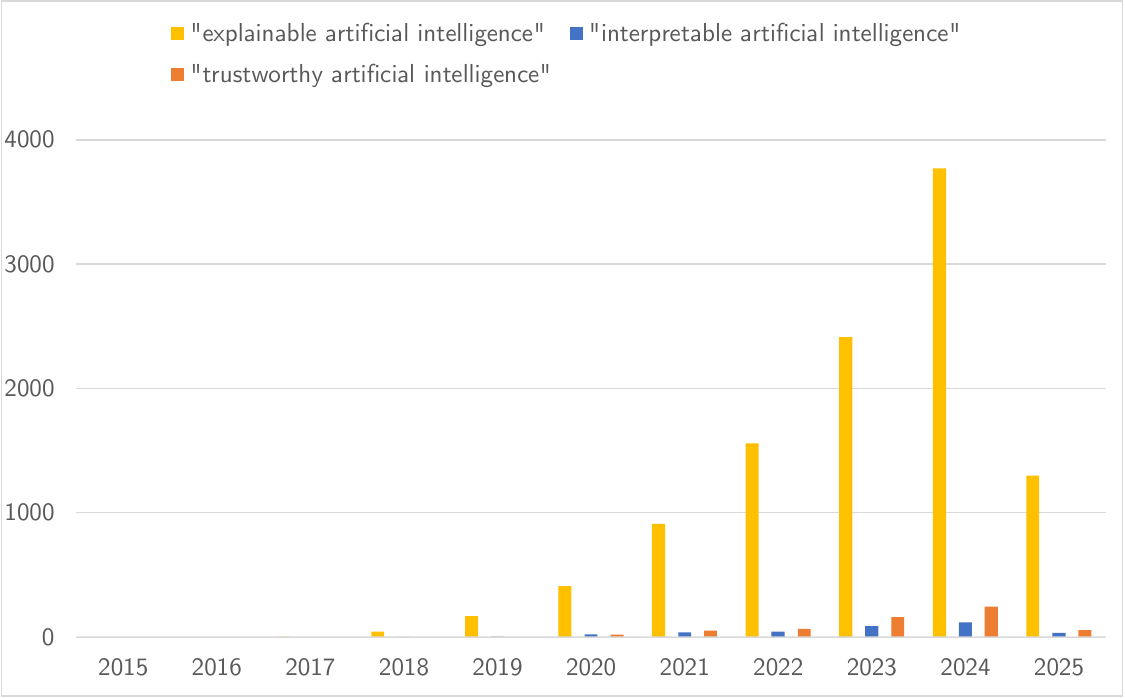}
\caption{Number of total publications and preprints referring to explainable, interpretable, and trustworthy AI since 2015. Data retrieved from Scopus (April 17th, 2025) by using the search terms: explainable artificial intelligence, interpretable artificial intelligence, trustworthy artificial intelligence}\label{publishing_trend}
\end{figure}

\noindent\textbf{Results of literature search.}
We present the results of our literature search conducted in Scopus. We limited the presentation to one database due to the issues described previously. 
We limited the search to years 2015-2025 and included preprints. 
First, we searched for papers on explainable, interpretable and trustworthy AI. There is a clear rising trend that can be associated with the rise of transformer-based models~\cite{lin_survey_2022}. Detailed results are presented in Figure \ref{publishing_trend}.

As depicted in Figure~\ref{publishing_trend}, significant progress has been made by researchers to expand the XAI toolbox~\cite{longo_explainable_2024} in the last years. However, interpretable and trustworthy AI constitute only a fraction of the publications.

We observe a significant dominance of publications on \textit{Explainable AI} over the others. Such a result is understandable, as this term has the broadest meaning and is the most popular in the research community. The narrowest term is \textit{Trustworthy}, a concept that has gained popularity recently, so the development is still in its early stages. \textit{Interpretable AI} can refer to self-explaining models, while there is a high demand for the interpretability of black-box models.

A similar situation is observed in the \textit{Web of Science} database. Publications related to "explainable artificial intelligence" clearly dominate: as of May 8, 2025, there were 1,963 publications recorded for 2024 (compared to 1,274 in 2023 and 934 in 2022). The keyword "trustworthy artificial intelligence" returned 76 publications in 2024, compared to 58 in 2023 and 13 in 2020 (with no records in earlier years). The keyword "interpretable artificial intelligence" returned 64 publications in 2024, compared to 51 in 2023, 12 in 2020, and one in 2019. The \textit{PubMed} database was not included in the analysis, as it would require manual verification of the retrieved results.
\begin{figure}[h]
\centering
\includegraphics[width=0.9\textwidth]{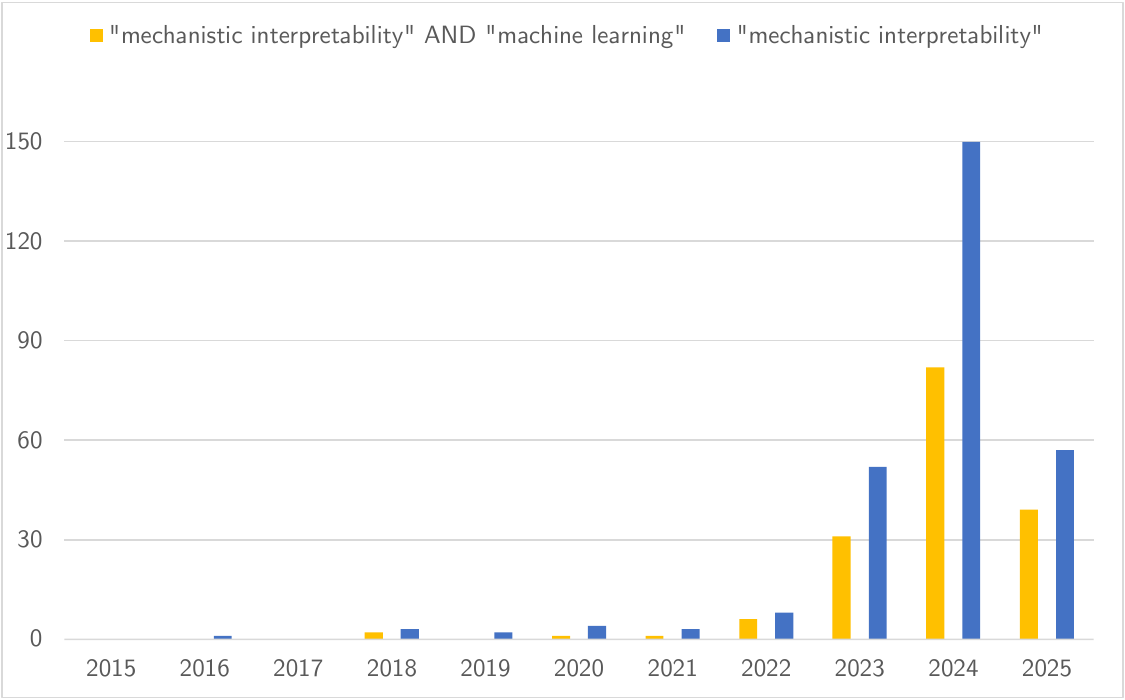}
\caption{Number of total publications and preprints referring to mechanistic interpretability since 2015. Data retrieved from Scopus (April 17th, 2025) by using the search term: "mechanistic interpretability" and "mechanistic interpretability"  AND  "machine learning"} 
\label{publishing_trend_mi}
\end{figure}

Figure~\ref{publishing_trend_mi} presents the number of publications referring to MI each year since 2015. The notable growth in publications since 2020 is related to the introduction of the term "mechanistic interpretability" in context of neural networks.
The difference in publication numbers for the two terms is a result of broadness. 
In the \textit{Web of Science}, the keyword "mechanistic interpretability" returned 23 publications in 2024 (compared to 9 in 2023 and 3 in 2022). The keywords "mechanistic interpretation" AND "machine learning" returned only five publications in 2024, two in 2023, and one publication each in 2022, 2021, 2020, and 2018.

Interest in mechanistic interpretability is currently limited compared to other areas of XAI research, as it requires significant expertise and effort to dissect complex models at a granular level. However, as depicted in Figure~\ref{publishing_trend_mi}, this field is rapidly gaining traction due to its promise of providing deeper insights into the inner workings of machine learning systems. Recent advancements demonstrate the feasibility of mechanistic interpretability, making it a promising avenue for addressing challenges in AI alignment, debugging, and ethical deployment. While still an emerging discipline, its ability to deliver detailed insights positions it as a critical component of the future of XAI research.


%
\section{Current state of MI research} \label{methods}
%
%
Mechanistic interpretability techniques encompass a diverse and evolving range of approaches aimed at understanding the inner workings of neural networks. These techniques range from narrow analyses, such as visualizing individual neurons and their activations, to approaches that examine the relationships between entire layers or components, such as circuits or attention heads. 
Based on the available literature, we have defined milestones for the development of MI area, presented in Table~\ref{tab:mi-literature}. The first column references the literature; the following columns contain the task the item concerns, a short description of the main contribution, and the model studied (where applicable). The last column indicates the year of the milestone.
%
\begin{table}[h] 
\small
\caption{Milestones in mechanistic interpretability.}
\label{tab:mi-literature}
\begin{tabularx}{\textwidth}{@{}llLl@{}} 
\toprule
\textbf{Reference} & \textbf{Task} & \textbf{Main contribution (model studied)} & \textbf{Year} \\
\midrule
\cite{olah_zoom_2020} & Circuit discovery & Introduction of circuits concept (InceptionV1) & 2020 \\
\cite{vig_causal_2020} & Circuit discovery & Application of CMA to circuits (GPT-2) & 2020 \\
\cite{elhage_mathematical_2021} & Reverse-engineering & Mathematical framework for transformers (toy transformer) & 2021 \\
\cite{geiger_causal_2021} & Hypothesis verification & Causal abstraction analysis (BiLSTM, BERT-based model) & 2020 \\
\cite{wang_interpretability_2022} & Circuit discovery & Introduction of path patching method (GPT-2 small) & 2022 \\
\cite{elhage_toy_2022} & Superposition study & Demonstration of superposition (toy models) & 2022 \\
\cite{sharkey_taking_2022} & Feature disentanglement & Recovery of ground truth features (small transformer) & 2022 \\
\cite{conmy_towards_2023} & Circuit discovery & Introduction of ACDC method (GPT-2 small) & 2023 \\
\cite{gurnee_finding_2023} & Locating features & Sparse probing technique (Pythia models) & 2023 \\
\cite{he_dictionary_2024} & Circuit discovery & Dictionary learning for circuits (Othello-GPT) & 2023 \\
\cite{geiger_causal_2024} & Other & Causal abstraction framework for MI & 2023 \\
\bottomrule
\end{tabularx}
\end{table}

Building on the taxonomy introduced in the previous section, we present the main techniques associated with each interpretability task. Each approach is presented with a conceptual description to explain its foundations, a mathematical framework where applicable to formalize its methodology, and multiple examples to illustrate its application in real-world neural networks.
\subsection{Feature localization} 
Feature localization refers to characterizing individual neurons. In MI, the common characteristic of a neuron is its \emph{importance} -- a measure of their contribution to the model's prediction.
Importance is usually measured as the activation or gradient value for a given input~\cite{dhamdhere_how_2018}. 
Studying individual neurons is the first step to reverse engineer a neural network, as they can be thought of as variables in the encoded algorithm. 

\subsubsection*{Probing} 
\begin{figure}[h]
    \centering
    \includegraphics{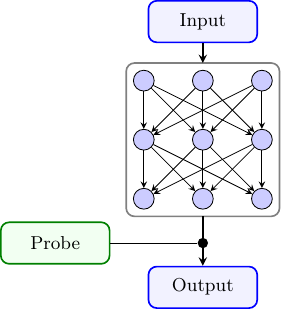}
    \caption{Illustration of probing. The probe does not interact with the neural network; rather it uses its representation to yield explanations. Note that probes can be applied at any layer of the network} 
    \label{probing}
\end{figure}

Probing is a technique derived from NLP to interpret and analyze the internal representations learned by models~\cite{alain_understanding_2017}. This method is based on the premise that neural networks do not acquire any new information during their forward passes but rather transform given inputs to learn meaningful representations. By training a simple classifier, called a \textit{probe}, on these generated representations, researchers can determine whether the network has captured any specific type of information~\cite{belinkov_probing_2022}. Figure~\ref{probing} illustrates the idea of probing.

Mathematically, probing can be framed as a supervised learning problem~\cite{hewitt_designing_2019}. Given a model \( f \) that generates hidden representations \( \mathbf{h}_i \in \mathbb{R}^d \) for each input token or sequence \( x_i \), the goal is to train a lightweight probe \( g \) that maps \( \mathbf{h}_i \) to a target property \( y_i \) (e.g., part-of-speech tag, dependency label). Formally, the probe learns a function \( g: \mathbb{R}^d \to \mathcal{Y} \), where \( \mathcal{Y} \) is the label space, by minimizing a loss function, \(\mathcal{L}(g(\mathbf{h}_i), y_i)\). The performance of the probe \( g \) on a held-out test set measures how well the representations \( \mathbf{h}_i \) encode the property \( y_i \)~\cite{alain_understanding_2017,belinkov_probing_2022}. Algorithm~\ref{algo_probing} presents
the pseudo-code of linear probing.

\begin{algorithm}[H]
\caption{Linear Probing of Internal Representations}
\label{algo_probing}
\begin{algorithmic}[1]
\Require Pretrained model \( f \), dataset \( D = \{(x_i, y_i)\}_{i=1}^n \), target layer \( L \), loss function \( \mathcal{L} \)
\State Initialize empty set \( \mathcal{H} \gets \emptyset \)
\For{each \( (x_i, y_i) \in D \)}
    \State \( \mathbf{h}_i \gets f_L(x_i) \) \Comment{Extract representation from layer \( L \)}
    \State Add \( (\mathbf{h}_i, y_i) \) to \( \mathcal{H} \)
\EndFor
\State Define probe \( g: \mathbb{R}^d \to \mathcal{Y} \) \Comment{e.g., linear classifier}
\State Train \( g \) on \( \mathcal{H} \) by minimizing \( \mathcal{L}(g(\mathbf{h}_i), y_i) \)
\State Evaluate \( g \) on a held-out test set
\State \Return Performance metrics (e.g., accuracy, F1)
\end{algorithmic}
\end{algorithm}
%
%

In language models, probing is used to assess whether the model has captured any linguistic knowledge. For instance, Conneau et al.~\cite{conneau_what_2018} introduce 10 probing tasks for single sentence embeddings and Lin et al.~\cite{lin_open_2019} probe BERT's~\cite{devlin_bert_2019} representations to identify structurally-defined elements. Gurnee and Tegmark~\cite{gurnee_language_2024} employ probing to identify neurons containing spatial and temporal features, and also provide evidence that validates the linear representation hypothesis (Section~\ref{disentanglement}).

In a mechanistic approach, probing is aimed at identifying whether and where the model encodes particular kinds of information, thereby complementing the mechanistic understanding~\cite{bereska_mechanistic_2024}. Probing can identify the roles of individual neurons and the features represented in the network, aiding in the localization of information~\cite{gurnee_finding_2023}. In particular \textbf{sparse probing}, which attempts to identify only a subset of neurons (or a single neuron) that is activated by a given input. This is achieved by limiting the classifier to $k$ non-zero coefficients, thus creating a \emph{$k$-sparse probe}. Sparse probing can be depicted as a neuron ranking problem -- scoring individual neurons based on their importance in a given task. It allows to filter out irrelevant neurons, highlighting the specific representations that encode linguistic or functional knowledge. However, due to the focus on sparsity, important neurons can be overlooked~\cite{antverg_pitfalls_2022,gurnee_finding_2023}. 

Gurnee et al.~\cite{gurnee_finding_2023} propose two techniques: adaptive tresholding, which constrains the value of $k$ through iterative retraining, and optimal sparse probing which employs a cutting plane algorithm for smaller values of $k$.
Chowdhury and Allan~\cite{chowdhury_understanding_2025} employ ridge regression-based probes to identify context neurons involved in the task of document ranking. The authors use Lasso regularization to enforce the sparsity in the classifiers. These probes are applied across the layers of the RankLlama model~\cite{ma_fine-tuning_2024}, revealing multiple statistical features captured along the layers of the network.

Other examples of probing include \textbf{edge probing} and \textbf{structural probing}, which are complementary methods used to evaluate linguistic knowledge encoded in language model representations, differing in focus and scope. Edge probing assesses whether local pairwise relationships between words or tokens (edges) in a linguistic structure are encoded within the representations of a model. Rather than classifying individual token properties, edge probes focus on the relationships or interactions between spans, such as syntactic dependencies or semantic roles~\cite{tenney_what_2019,choudhury_implications_2023}. 

In contrast, structural probing evaluates the encoding of global hierarchical structures, such as full syntactic dependency trees or semantic frames. Structural probes examine how well linguistic structures are embedded in model representations, capturing the relationships between all elements in a sentence. For instance, Hewitt and Manning~\cite{hewitt_designing_2019} present a structural probe to identify syntax trees embedded within ELMo~\cite{peters_deep_2018} and BERT~\cite{devlin_bert_2019} representations by predicting distances or reconstructing hierarchical tree structures. Structural probing provides insights into the broader organization of linguistic information encoded in embeddings. Yet, the self-supervised nature of structural probes raises doubts whether the probe actually captures any knowledge of language models~\cite{farquhar_challenges_2023}.

\subsubsection*{Lenses}
In contrast to probing, lenses do not focus solely on observing a specific state of the model. Rather, they map the representations from intermediate layers to the model's vocabulary distribution, without passing them through the remaining layers of the model. By revealing how the model's predictions develop across the network, this approach allows to investigate what linguistic or semantic information is encoded at different stages of the model's computation~\cite{elhage_mathematical_2021,geva_transformer_2021}. Formally, a lens maps the hidden state $h^l$ at layer $l$ to the vocabulary space by multiplying it with the output embedding matrix, $\boldsymbol{W_{out}}$, yielding the vocabulary logits, $z_l$~\cite{rai_practical_2025}:
\begin{equation}
\boldsymbol{z_l} = \boldsymbol{W_{out} h^l}
\nonumber
\end{equation}
The \textbf{logit lens}, first introduced by nostalgebraist~\cite{nostalgebraist_interpreting_2020} in 2020, was quickly adopted by researchers and applied in various transformer architectures~\cite{luo_understanding_2024}. Algorithm~\ref{algo_logit_lens} presents
the pseudo-code of the lens. Early works focused on analyzing only specific elements of transformer's architecture: Geva et al.~\cite{geva_transformer_2021,geva_transformer_2022} analyze how feed-forward layers update token representations, enabling these representations to be interpreted at any stage of the model's computation, and Sakarvadia et al.~\cite{sakarvadia_memory_2024} map attention heads to the vocabulary space and identify a mechanism responsible for retrieving memories. Based on the attention module's memory, the lens has also been employed to model the information flow in GPT-2~\cite{katz_visit_2023}. Dar et al.~\cite{dar_analyzing_2023} extend upon the logit lens by taking into account all weights of a model and offer a framework for interpreting model's parameters in isolation. Merullo and Eickhoff~\cite{merullo_language_2024} employ the logit lens in an in-context learning setting, uncovering a mechanism that implements simple vector arithmetic. A different application of the lens was demonstrated by Wendler et al.~\cite{wendler_llamas_2024}, who investigated whether the representations of the Llama-2 model~\cite{touvron_llama_2023} consistently use English as their pivot language regardless of the prompt. 

\begin{algorithm}[H]
\caption{Logit Lens}
\label{algo_logit_lens}
\begin{algorithmic}[1]
\Require Model \( f \), output embedding matrix \( \boldsymbol{W_{out}} \), input sequence \( x \), layer indices \( \mathcal{L} = \{1, 2, \dots, L\} \)
\State Tokenize input: \( x \rightarrow (x_1, x_2, \dots, x_T) \)
\State Pass input through model to extract hidden states \( \{h^l\}_{l \in \mathcal{L}} \)
\For{each layer \( l \in \mathcal{L} \)}
    \State Compute vocabulary logits: \( \boldsymbol{z_l} = \boldsymbol{W_{out} h^l} \)
    \State Store or visualize top-k predictions from \( \boldsymbol{z_l} \)
\EndFor
\State \Return Layer-wise vocabulary predictions \( \{\boldsymbol{z_l}\}_{l \in \mathcal{L}} \)
\end{algorithmic}
\end{algorithm}

An extension of the logit lens is the \textbf{DecoderLens}, which overcomes its limitations in interpreting representations within encoder-decoder models. Langedijk et al.~\cite{langedijk_decoderlens_2024} propose enabling the decoder to cross-attend to intermediate encoder layer representations, rather than relying solely on the final encoder output, providing richer insights into the model's internal workings.

Other techniques include the \textbf{linear lens}, which questions the idea of mapping hidden representations directly to the output distribution. Din et al.~\cite{din_jump_2024} cast hidden representations between layers by fitting a linear regression from source layer $l_i$ to target layer $l_{i+1}$. Similarly, the \textbf{tuned lens}~\cite{belrose_eliciting_2023} trains affine transformations in order to \emph{translate} representations from an internal layer to the final layer. The \textbf{future lens} investigates whether future tokens can be predicted from hidden states. To achieve this, Pal et al.~\cite{pal_future_2023} train a linear model $f_{\theta}$ transforming a hidden encoding at layer $l$, $h^l_T$, to a future hidden state at layer $L$, $h^l_{T+1}$. A further example is the \textbf{JailbreakLens}, an interpretation framework introduced by He et al.~\cite{he_jailbreaklens_2024}, which investigates how jailbreak impacts LLM's representations. Lenses have also been applied beyond LLMs; for instance, the \textbf{diffusion lens} employed in text-to-image diffusion models~\cite{toker_diffusion_2024} and the \textbf{SemanticLens} -- an universal explanation tool which maps neural network's components into a semantically structured space~\cite{dreyer_mechanistic_2025}.

\subsection{Circuit discovery} 

Circuit discovery aims to identify causal graphs in neural networks -- structured subnetworks of neurons and connections that encode specific computational functions. This process validates the algorithmic nature of deep AI models, thereby reinforcing the idea of mechanistic interpretability and moving beyond black-box explanations. Empirical studies have demonstrated the emergence of structured circuits across neural networks, both in small toy models~\cite{nanda_progress_2023,zhong_clock_2023,kitouni_neurons_2024} and large-scale real-world networks~\cite{lieberum_does_2023,wang_interpretability_2022}. These findings reinforce the view that neural networks, despite their complexity, operate through interpretable and decomposable computational structures rather than relying solely on distributed, uninterpretable representations.

\begin{figure}[h]
    \centering
    \includegraphics{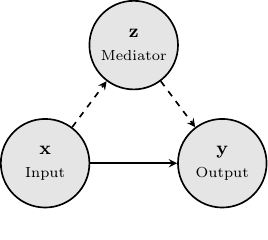}
    \caption{Causal mediation analysis} 
    \label{cma}
\end{figure}

The main framework for circuit discovery is based on Causal Mediation Analysis (CMA). CMA is a method derived from causal inference used to study systems of dependencies, typically represented by directed acyclic graphs~\cite{pearl_causal_1995}. It builds upon the causal framework by including \emph{mediators} in its analyses -- intermediate elements in a system that influence the final outcome. In neural networks, causal mediation analysis can be used to investigate circuits by treating internal model components (e.g., neurons) as mediators between inputs and outputs~\cite{vig_causal_2020}. Figure \ref{cma} depicts a causal graph with a mediator.

The core method derived from CMA in mechanistic interpretability is \textbf{patching} (also known as interchange intervention~\cite{geiger_causal_2024}) -- a method employed to study and understand the internal computations of a machine learning model by modifying -- patching -- its internal components during inference. The objective is to identify specific components of the model (e.g., particular neurons, attention heads, layers) that are causally responsible for certain behaviors or outputs. Patching stands as a fundamental approach in MI, as it is by definition not limited to studying only the input-output relations but emphasizes the importance of internal components~\cite{vig_causal_2020}. Patching aims to uncover computational circuits by intervening in model components and observing the resulting changes in output. This technique views models as causal graphs in which internal components (such as neurons) form causal pathways connecting inputs to outputs \cite{finlayson_causal_2021,geiger_causal_2021,stolfo_mechanistic_2023,vig_causal_2020}. 

\subsubsection*{Activation patching} Activation patching stands as the main technique derived from causal interference. The activation-based approach focuses on modifying or replacing the activations of neurons within a neural network model. It's primary use is to understand the role of specific neurons or layers in the model's decision-making process by observing how the changes in activations impact the model's output~\cite{bereska_mechanistic_2024}. Activation patching follows a standard workflow to identify crucial activations in neural networks~\cite{heimersheim_how_2024}: 
\begin{enumerate}
    \item Clean run -- the model processes a clean input, generating a correct prediction, and caches the activations.
    \item Corrupted run -- the model processes a corrupted input, so that the prediction is impaired.
    \item Patched run -- the model processes a corrupted input, but selected activations are substituted with those from the clean run.
\end{enumerate}
Algorithm~\ref{algo_patching} presents the pseudo-code of activation patching.
\begin{algorithm}[H]
\caption{Activation Patching}
\label{algo_patching}
\begin{algorithmic}[1]
\Require Model \( f \), clean input \( x_{\text{clean}} \), corrupted input \( x_{\text{corrupt}} \), layers \( \mathcal{L} \), positions \( \mathcal{T} \), selected components \( \mathcal{C}^l_t\)
\State Run forward pass on \( x_{\text{clean}} \) to obtain activations \( h^l_{\text{clean}}[t] \)
\State Run forward pass on \( x_{\text{corrupt}} \) to obtain activations \( h^l_{\text{corrupt}}[t] \)
\For{each component \( c \in \mathcal{C}^l_{t} \)}
    \State In the corrupted run, replace \( h^l_{\text{corrupt}}[t] \gets h^l_{\text{clean}}[t] \)
    \State Continue the forward pass from layer \( l+1 \) onward using patched activations
    \State Record model output after patching (e.g., logits or probabilities)
    \State Compare patched output with clean and corrupted outputs
\EndFor
\State \Return Degree of restoration (e.g., logit difference, KL divergence)
\end{algorithmic}
\end{algorithm}
Based on the performance of the model during the patched run, researchers can determine the significance of the replaced components, narrowing down a computational circuit. Conmy et al.~\cite{conmy_towards_2023} propose a method for automating the workflow -- ACDC (Automated Circuit DisCovery), an iterative algorithm that evaluates and patches activations at each node so that the performance of the model remains unaffected. However, in both manual and automated approaches, the substitutions of activations are performed iteratively, resulting in a substantial computational cost of this technique. Other automation techniques include more advanced, more accurate, and scalable ACDC, named ACDC++~\cite{syed_attribution_2023} and Contextual
Decomposition for Transformers (CD-T)~\cite{hsu_efficient_2024}. 

Activation patching is employed in identifying critical activations for storing and processing data; for instance, Stolfo et al.~\cite{stolfo_mechanistic_2023} find that language models process arithmetic information by transmitting relevant data from early to late layers using the attention mechanism. This information is then processed by a set of MLP modules, which generate the final result-related information incorporated into the residual stream. Meng et al.~\cite{meng_locating_2022} employ activation patching to investigate the storage and recall of factual associations within autoregressive transformer language models. Other researchers investigate circuits in tasks such as syntactic agreement~\cite{finlayson_causal_2021} or multiple-choice question answering~\cite{lieberum_does_2023}. Furthermore, by employing activation patching, Todd et al.~\cite{todd_function_2024} find a key mechanism of in-context learning: function vectors, which trigger the execution of specific procedures in a language model. Marks and Tegmark~\cite{marks_geometry_2024} employ activation patching to locate the representations of truth in the LLaMA-2 model, and  further analyze these hidden states through visualization and probing techniques. Lan et al.~\cite{lan_towards_2024} investigate motifs through activation patching and demonstrate a circuit responsible for sequence continuation tasks. Monea et al.~\cite{monea_glitch_2024} enhance activation patching by focusing on corrupting tokens rather than embeddings. Aside language models, activation patching was employed by Palit et al.~\cite{palit_towards_2023} to study visual question answering in BLIP~\cite{li_blip_2022}. Their results suggest that the vision modality is located in the final layers of the model. 

Although the idea of localizing important components in neural networks through interventions seems sensible, researchers have questioned the faithfulness of such interpretations. Although the interventions may affect the responses of the model, it is not sufficient to clearly establish the component as meaningful~\cite{wang_does_2024}. Hase et al.~\cite{hase_does_2023} question the reliability of mechanistic understanding in locating models' behaviors. Current research aims to increase both the interpretability and the faithfulness of circuits by optimizing intervention techniques~\cite{stoehr_activation_2024,wang_does_2024}.

\subsubsection*{Attribution patching} Attribution patching emerged as a response to the computational complexity of the standard procedure of activation patching~\cite{nanda_attribution_2023}. Instead of performing a forward pass for each substitution, this approach uses a gradient-based approximation to find a linear estimate between the corrupted and clean pass. This allows to find the circuit in only two forward passes and one backward pass, simultaneously reducing the need for human supervision during the process~\cite{ferrando_information_2024}. Syed et al.~\cite{syed_attribution_2023} extend this approach by proposing \textbf{edge attribution patching}, an automated circuit discovery algorithm that obtains the most important edges in a given task. Hanna et al.~\cite{hanna_have_2024} further build upon edge attribution patching by integrating gradients to increase the faithfulness of the discovered circuits.

\subsubsection*{Path patching} Path patching constrains the interventions to a specific path in the model, leaving the rest of the network untouched. The goal is to determine whether the patched path causally contributes to a specific behavior or output. In contrast to other patching method, this method allows to investigate the impacts of components in a network on each other~\cite{hanna_how_2023}. Wang et al.~\cite{wang_interpretability_2022} show its application in discovering a circuit in GPT-2 small. Goldowsky-Dill et al.~\cite{goldowsky-dill_localizing_2023} generalize the idea for any number of paths, modeling the network as an arbitrary computational graph.


\subsubsection*{Ablation study} Ablation, also known as knockout, is another technique derived from causal interference. Although both ablation studies and patching involve the manipulation of a system's components, they differ in their specific approaches and objectives. Ablation is primarily focused on removing or disabling certain elements within the system to observe the resulting changes in behavior, thus identifying which components are essential for the functionality of the system~\cite{meyes_ablation_2019}. In contrast to patching, ablation does not seek to understand the causal relationships between components; rather their causal importance~\cite{li_circuit_2024,lan_towards_2024}. The removal of components is usually performed via \emph{zero ablation}, which sets their values to zero, and \emph{mean ablation}, which replaces their values with the average values across a training distribution. However, zero ablation often results in noisy results, as setting components to zero may introduce artificial disruptions that do not naturally occur in the model~\cite{wang_interpretability_2022}. A novel approach is \emph{optimal ablation}, proposed by Li and Janson~\cite{li_optimal_2024}, that sets the value to constant that yields the minimal loss of the ablated model. 

Ablation has been applied to various components of neural networks. Ghorbani and Zou~\cite{ghorbani_neuron_2020} introduced Neruon Shapley, a framework employing zero ablation to identify the crucial neurons. In their work, they quantify the importance score of neurons based on their estimated Shapley value, taking into account interaction between components (e.g., if two neurons are both required to improve the performance of the model). Alternatively, Ollson et al.~\cite{olsson_-context_2022} study the causal importance of induction heads in in-context learning. Geva et al.~\cite{geva_dissecting_2023} and Fierro et al.~\cite{fierro_how_2025} knock out attention weights to investigate how GPT models extract factual knowledge. Li et al.~\cite{li_circuit_2024} focus on disabling edges responsible for bad behaviors in models (e.g., hallucinations) using \textbf{targeted edge ablation}, and García-Carrasco et al.~\cite{garcia-carrasco_detecting_2024} combine ablation with the logit lens to detect vulnerabilities in GPT-2 Small~\cite{radford_language_2019}.

Another example of ablation study is \textbf{causal scrubbing}, a generalized approach that aims to test hypotheses about why does a model behave in a specific way. In contrast to other causal approaches, scrubbing begins with establishing a hypothesis about the circuits. The algorithm introduced by Chan et al.~\cite{chan_causal_2022} is as follows:
\begin{enumerate}
    \item Define the components and pathways that the hypothesis claims are responsible for a mechanism.
    \item Intervene on parts of the model that are not specified by the hypothesis.
    \item Evaluate the output of the model -- if the hypothesis is correct, it should remain unchanged.
\end{enumerate}

In~\cite{chan_causal_2022}, the interventions were applied using \emph{resampling ablation}, which replaces the values of the components using any value they had on other inputs. Due to the restriction on the intervention space, causal scrubbing performs less interventions than other methods. However, it relies heavy on the hypotheses and can misidentify correlated features~\cite{jenner_comparison_2023}. 

\subsection{Feature disentanglement} \label{disentanglement}
Feature disentanglement methods aim to resolve the issue of polysemanticity, which represents a significant challenge to mechanistic interpretability. Ideally, a neuron would respond to a single feature, making the network monosemantic and easier to interpret. However, in practice, neurons are often observed to be polysemantic, meaning that they respond to multiple, often unrelated concepts~\cite{goh_multimodal_2021,olah_feature_2017}. The primary reason for polysemanticity is the substantial number of features that the network attempts to accommodate within its limited number of neurons~\cite{olah_zoom_2020}. 
Furthermore, Lecomte et al.~\cite{lecomte_what_2024} highlight the phenomenon of \emph{incidental polysemanticity}, where a network, despite having enough neurons, still overlaps representations of distinct features. 

\begin{figure}[h]
    \centering
    \includegraphics{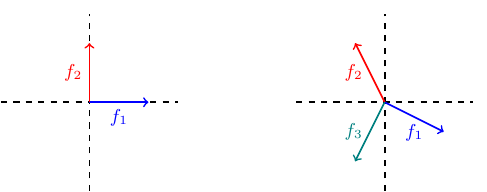}
    \caption{Superposition of features in two-dimensional space. Left: feature directions are orthogonal, no polysemanticity. Right: Overlapping (almost orthogonal) feature directions, polysemanticity} 
    \label{superposition}
\end{figure}

The \textit{superposition hypothesis} provides a theoretical framework to explain this behavior, proposing that polysemanticity results from the \emph{compression} of features into a limited representational space. he features are encoded not only in the $n$-dimensional activation space of a network but also in the almost orthogonal directions, as illustrated in Figure~\ref{superposition}. Elhage et al.~\cite{elhage_toy_2022} demonstrate that superposition emerges naturally in neural networks. Further, Scherlis et al.~\cite{scherlis_polysemanticity_2023} show that polysemanticity depends on the allocated importance of features, i.e., important features are encoded orthogonally to other features. Other explanations behind polysemanticity include redundancy introduced during training (e.g., due to random dropout), or correlations in natural data that encourage neurons to share features~\cite{chan_superposition_2024}.

The goal of feature disentanglement is to enumerate all the features encoded by the network, thereby providing a clearer understanding of its internal representations. Disentanglement can be applied ad-hoc, by designing models without superposition~\cite{elhage_toy_2022}, and post-hoc, by employing sparse coding to describe how features are represented in models with superposition~\cite{sharkley_current_2022}. 

\subsubsection*{Monosemanticity in models} The motivation behind designing models with monosemantic neurons is their increased interpretability, as such neurons can be easily mapped to features. One possibility is the use of linear models, which do not exhibit superposition~\cite{elhage_toy_2022}. In non-linear networks, superposition can be reduced by decreasing the sparsity of features. However, this forces models to learn only the top $n$ important features. Jernym et al.~\cite{jermyn_engineering_2022} engineer monosemanticity in toy models and show that increasing the number of neurons per layer reduces polysemanticity, yet leads to increased computational cost. The benefits of monosemantic models are a subject of ongoing debate within the research community; Wang et al.~\cite{wang_learning_2024} demonstrate that decreasing monosemanticity benefits model's performance, whereas Yan et al.~\cite{yan_encourage_2024} show a positive correlation between monosemanticity and model's capacity. The question whether monosemanticity should be inhibited or encouraged remains unsettled.

\subsubsection*{Sparse autoencoders} 

Following the superposition hypothesis, activations can be decomposed as a linear combination of features (directions in the activation space)~\cite{elhage_toy_2022,park_linear_2024}. Based on this premise, a natural solution to finding the single features encoded by a network is the implementation of sparse coding~\cite{olshausen_sparse_1997}, especially sparse autoencoders (SAEs). 
Also known as sparse dictionary learning, the problem can be framed mathematically as follows: given an input dataset, $X=[\boldsymbol{x_1},...,\boldsymbol{x_i}]$, $\boldsymbol{x_i} \in \mathbb{R}^m$, and a positive integer $n$, find a dictionary $\boldsymbol{D} \in \mathbb{R}^{m \times n}$, that allows for the sparsest representation $R=[\boldsymbol{r_1},...,\boldsymbol{r_i}]$, $\boldsymbol{r_i} \in \mathbb{R}^n$ so that $\|\boldsymbol{DR}-\boldsymbol{x}\|_2$ is minimized~\cite{tillmann_computational_2015}. 

In other words, SAEs transform an entangled space into a sparse representation, enabling independent analysis of features. Such decomposition offers a great opportunity for interpretability, as it allows for analyzing the network's components in isolation. SAEs have been successfully employed for word embedding representations visualization in LLMs~\cite{arora_linear_2018,panigrahi_word2sense_2019,subramanian_spine_2018,yun_transformer_2021}, and more recently to extract features from LLMs~\cite{he_llama_2024,huben_sparse_2024,deng_measuring_2023}. In the context of mechanistic interpretability, SAEs have been integrated as part of circuit discovery pipelines~\cite{ge_automatically_2024,he_dictionary_2024,marks_sparse_2024,oneill_sparse_2024}.

SAEs are not the only approach to feature disentanglement. O'Mahony et al.~\cite{omahony_disentangling_2023} proposed disentangling polysemantic neurons into concept vectors, which are yielded from intermediate representations of images that have the highest activation values for a neuron. Dunefsky et al.~\cite{dunefsky_transcoders_2024} studied \textit{transcoders}, a variant of SAEs which aims to reconstruct the original layer output, and demonstrated its use in circuit discovery.
%
5
\section{Challenges and opportunities} \label{future}
Despite its promises, mechanistic interpretability still encounters numerous challenges. In this section, we address the major issues related to MI and outline some potential opportunities and applications.
\subsection{Challenges}
Mechanistic interpretability promises to fully uncover the computations of neural networks. However, many researchers have posed the question whether achieving this ambitious goal is truly possible. 
The current research is limited to dissecting toy models or small fragments of larger neural networks. The narrow scope of MI pursuits may lead to "cherry-picked" achievements, with methods lacking broader evaluation~\cite{rauker_toward_2023}. We outline the major challenges facing the field of mechanistic interpretability, followed by a detailed description of the key problem areas.

The following challenges have been identified:
\begin{itemize}
    \item \textit{Superposition}: Individual neurons often encode multiple unrelated concepts, leading to multiple possible interpretations during mechanistic analysis~\cite{elhage_toy_2022}.
    \item \textit{Spurious correlations}: MI methods can uncover mechanisms that rely on coincidental or non-causal patterns in the data, potentially leading to misleading interpretations~\cite{nanda_progress_2023}.
    \item \textit{Scalability}: MI methods are largely limited to small or simplified models; applying them to real-world problems remains technically infeasible and labor-intensive~\cite{cammarata_thread_2020,nanda_progress_2023}.
    \item \textit{Evaluation problems}: The absence of accepted metrics for evaluating MI methods in terms of the faithfulness, completeness, or usefulness of interpretations undermines the reliability of mechanistic explanations~\cite{hanna_have_2024,meloux_everything_2025}.
    \item \textit{Deficit of automation}: Circuit discovery methods require extensive manual work in both circuit analysis and intervention design, limiting reproducibility and scalability~\cite{conmy_towards_2023}.
    \item \textit{Semantic drift}: The meanings of internal representations can shift across network layers or training checkpoints, challenging longitudinal analyses~\cite{olah_zoom_2020,elhage_mathematical_2021}.
    \item \textit{Pitfalls of intervention}: Every intervention may potentially influence the model behavior, making the causal importance of components uncertain and raising questions about the robustness of findings~\cite{wang_does_2024}.
  \end{itemize}  

\noindent\textbf{Spurious explanations vs. genuine understanding.} A key challenge in mechanistic interpretability is distinguishing between explanations that accurately reflect the model’s internal computations and those that are merely artifacts of human-imposed narratives. 
Cherry-picked results are not enough to demonstrate the understanding of a neuron. Currently, with no unified evaluation techniques, results from MI research are based on human interpretations. This may lead to false assumptions about a model’s reasoning process, undermining the reliability of MI studies~\cite{yang_common_2022}. Sharkey et al.~\cite{sharkey_open_2025} and Rauker et al.~\cite{rauker_toward_2023} emphasize the importance of validating hypotheses about models' behavior and the development of interpretability benchmarks. The comparison and validation of MI pursuits is only possible through the utilization of standardized evaluation techniques.

Faithful interpretations should capture the true causal mechanisms driving model behavior rather than providing post hoc justifications that appear convincing but lack real explanatory power. Recent work by Meloux et al.~\cite{meloux_everything_2025} tackles the issue of identifiability of MI explanations by assessing both the uniqueness of discovered circuits and the algorithms they encompass. Their experiments in toy models show that multiple interpretations can be assigned to a given circuit. \newline

\noindent\textbf{Practical deployment.}
Mechanistic interpretability has seen only a limited application in non-research contexts. The following factors have contributed to this limited utilization: the gradual emergence of MI into the mainstream~\cite{saphra_mechanistic_2024}, MI research being limited to a family of models~\cite{sharkey_open_2025}, and the continuous need for human input and validation~\cite{bereska_mechanistic_2024}. Moreover, the findings from studies conducted on toy models may not be directly applicable to large-scale neural networks in the real world.
Despite the advancement in the automation of MI workflows and the scaling up of techniques, further research is required to fully explore the potential of this approach to interpretability. \newline

\noindent\textbf{Universality.}
Interpretability is a difficult and laborious process, requiring careful analysis from a human specialist. Each individual case requires independent interpretation of the results. However, knowledge transfer between models and architectures could greatly accelerate advances in mechanistic interpretability. If the universality hypothesis~\cite{chughtai_toy_2023,merullo_circuit_2024} is proven to be true, similar structures could be identified in multiple models. This would allow researchers to apply insights from one study to another.
However, the hypothesis has only been proven in toy models and for specific tasks. There is no guarantee that the results will be applicable to larger, more advanced architectures~\cite{friedman_interpretability_2024}. Work in this field remains an open challenge in MI. \newline

\noindent\textbf{Internal understanding.}
Mechanistic interpretability requires researchers to navigate the intricate landscape of a model’s internal components and representations. However, the task of understanding and translating these inner workings into human-understandable processes is complicated by challenges such as superposition, semantic drift, and the effects of experimental interventions. Superposition entangles multiple features within the same neurons or activations, requiring careful disentanglement to isolate and interpret meaningful representations. Semantic drift introduces additional uncertainty, as the meaning of features and concepts can shift across different layers of the network. Moreover, the very act of intervening in a model—whether by modifying activations or probing circuits—can itself alter the system’s behavior, potentially confounding the interpretation. 
Combined with cherry-picking and confirmation bias, these complexities lead to an incomplete or overly optimistic picture of a model's interpretability~\cite{burns_discovering_2023}. As a result, ensuring rigor, transparency, and reproducibility is essential for avoiding biased conclusions in MI research. Recognizing and addressing these methodological pitfalls is critical to advancing a more reliable and scientific understanding of machine learning systems.


\subsection{Opportunities}

Despite the multiple challenges present in MI, it is still a promising area of research. The possibilities in context of AI safety convince researchers to continue work in this field. We outline the key opportunities, followed by a detailed description.

Potential advantages of applying MI are:
\begin{itemize}
    \item \textit{Debugging}: MI enables the identification of faulty reasoning or unexpected model behavior by tracing decisions to specific circuits or internal mechanisms, helping to fix or retrain models more effectively~\cite{wang_interpretability_2022,olah_zoom_2020}.
    \item \textit{Data privacy}: By understanding how models store and access information internally, MI can help reveal whether sensitive training data is memorized or indirectly encoded, aiding privacy audits~\cite{carlini_extracting_2023}.
    \item \textit{Robustness}: MI can highlight the features and mechanisms that models rely on, exposing vulnerabilities to adversarial inputs or distribution shifts and guiding strategies to make models more robust~\cite{ilyas_adversarial_2019,burns_discovering_2023}.
    \item \textit{Control}: Gaining mechanistic insight opens the possibility of directly editing model behavior by modifying or steering internal components, supporting efforts in fine-tuning and alignment~\cite{sharkey_open_2025,meng_mass-editing_2023}.
    \item \textit{Human-in-the-loop}: MI supports collaborative AI development by allowing humans to interpret and interact with models' decision processes, enabling feedback loops for supervision and correction~\cite{chung_human_2021}.
    \item \textit{Trustworthiness}: By revealing the internal logic behind model outputs, MI can provide evidence of consistent, rational behavior, supporting confidence in AI systems used in high-stakes domains.
\end{itemize}

\noindent\textbf{AI safety.} Mechanistic interpretability is focused on establishing the roles of components in neural networks -- neurons, attention heads, or whole layers. MI does not only rely on the input-output relations, but also studies the causal dependencies in between, thus dissecting the model's decision-making process step by step. This makes MI methods valuable tools for locating features -- both valuable, like for instance, factual information~\cite{meng_locating_2022,geva_dissecting_2023,fierro_how_2025} or truth representations~\cite{marks_enhancing_2024}, and dangerous, such as hallucinations~\cite{li_circuit_2024,garcia-carrasco_detecting_2024,yu_mechanistic_2024} or toxic neurons~\cite{yang_beyond_2024}. Li et al.~\cite{li_circuit-tuning_2025} further extend this idea by locating and removing redundant parameters, demonstrating that MI insights can be applied to design more robust and efficient models.

Mechanistic interpretability seeks to elucidate the inner workings of neural networks in order to increase their transparency and trustworthiness. MI has been applied in the task of knowledge localization, debugging, and removal of harmful components. \newline

\noindent\textbf{Algorithmic understanding of neural networks.}
Mechanistic interpretability aims to identify \textit{how} information is processed within neural networks. This involves reverse-engineering internal circuits and structures, and expressing them in human-interpretable terms. MI offers a more granular understanding of a model’s decision-making processes by recovering routines that resemble hand-engineered algorithms.
Achieving algorithmic understanding holds significant promise for improving our ability to reason about model generalization, identify failure modes, and precisely intervene on internal behavior. For instance, Chen et al.~\cite{chen_axiomatic_2024} identified a circuit responsible for computing document relevance in the context of document ranking. \newline

\noindent\textbf{Real-world applications.} 
While much of the foundational work in MI has focused on general-purpose AI systems, recent studies have explored its applications in specific domains. For instance, in finance: Golgoon et al.~\cite{golgoon_mechanistic_2024} applied mechanistic interpretability techniques to GPT-2 Small, aiming to uncover how the model identifies potential violations of Fair Lending laws. 
Medicine has seen no implementation of mechanistic interpretability thus far. Nevertheless, research has highlighted the necessity for transparent and interpretable AI systems; therefore MI appears to be an area of considerable potential~\cite{s_band_application_2023,wang_framework_2024}.

\section{Conclusions} \label{conclusion}

Mechanistic interpretability has emerged as a promising research program within the broader landscape of explainable artificial intelligence. In recent years, deep learning has achieved substantial progress across diverse domains, including natural language processing, computer vision, and scientific discovery. Despite these advances, a fundamental challenge remains: understanding the mechanisms by which these models generate their predictions. MI addresses this issue by aiming to uncover the underlying computations within neural networks, providing insight into the specific mechanisms that give rise to observed behaviors. This paper sought to provide a comprehensive entry point into the field of MI.
We introduced MI by providing its historical and scientific context (Section~\ref{what_is_mi}). We presented a unified taxonomy of MI approaches (Section~\ref{what_is_mi}) and explored in detail multiple techniques, provided examples and pseudo-code for representative MI methods (Section~\ref{methods}). Our literature review demonstrated the rapidly growing interest in the field (Section~\ref{literature}).

One of the central promises of mechanistic interpretability lies in its potential to move beyond post-hoc correlations and surface-level attributions -- common in other XAI approaches -- toward causal, structural, and component-level understanding of neural network behavior. MI aspires to uncover algorithmic structures that are implemented within the weights and activations of trained models. By reverse-engineering these internal mechanisms, researchers can gain insight into the representations learned by the model, the decision-making procedures it implements, and the ways in which it generalizes (or fails to generalize) across tasks and inputs.

A strong mechanistic understanding is advantageous in a number of ways. MI could improve model debugging and failure mode analysis by identifying the specific pathways that contribute to undesirable behaviors. It may enable model editing -- the ability to intervene on internal components to change a model's outputs in a controlled and interpretable way. 
A future subject to be studied is the transfer of circuits between models, addressing whether and how discovered functional circuits from one model can be used in another.
An additional area worth exploring is the representation and integration of different modalities in mechanistic circuits.
Furthermore, MI holds the promise of supporting safety and alignment efforts, particularly in the context of powerful foundation models, where understanding the internal dynamics of goal representation, deception, or power-seeking behavior may be critical. MI could also contribute to the development of more scientific AI, with machine learning systems serving not only as effective tools but also as objects of study in their own right -- similar to the manner in which the study of natural intelligence is conducted in neuroscience and cognitive science.

At the same time, the field of MI is still in its early stages and many challenges remain. Among these are scaling interpretability techniques to the complexity of frontier models, developing principled evaluation frameworks for mechanistic claims, and determining the limits of human comprehensibility when confronted with distributed representations and high-dimensional feature spaces. There are also questions regarding the degree to which mechanistic insights transfer across architectures, training regimes, or tasks. 

We view this article as a foundational step toward consolidating the field of mechanistic interpretability. By providing a coherent definition, a taxonomy of techniques, and a structured overview of the current research landscape, we hope to support further theoretical development, methodological refinement, and interdisciplinary dialogue. The long-term vision of MI is ambitious: to build a science of deep learning systems that enables not only prediction and control but also understanding.


\bibliography{references}

\end{document}